\documentclass[conference]{IEEEtran}
\IEEEoverridecommandlockouts
\usepackage{cite}
\usepackage{amsmath,amssymb,amsfonts}
\usepackage{algorithmic}
\usepackage{graphicx}
\usepackage{textcomp}
\usepackage{booktabs,comment}
\usepackage[pagebackref=true,breaklinks=true,colorlinks,bookmarks=false]{hyperref}
\usepackage[dvipsnames]{xcolor} 
\usepackage[caption=false]{subfig}

\def\BibTeX{{\rm B\kern-.05em{\sc i\kern-.025em b}\kern-.08em
    T\kern-.1667em\lower.7ex\hbox{E}\kern-.125emX}}
\begin{document}

\title{End-to-end optimized image compression for multiple machine tasks\\
\author{Lahiru D. Chamain$^{\dag\ast}$\thanks{Work carried out while Lahiru D. Chamain was interning at the InterDigital AI Lab.}
	\qquad Fabien Racapé$^{\dag}$\qquad   Jean Bégaint$^{\dag}$\qquad   Akshay Pushparaja$^{\dag}$\qquad   Simon Feltman$^{\dag}$\\
	\begin{minipage}{0.95\textwidth}
		\small
		\centering
		\begin{tabular}{cc}
			$^{\dag}$InterDigital \-- AI Lab    & $^{\ast}$University of California, Davis \\
			4410 El Camino Real                 & 1 Shields Ave \\
			Los Altos, CA, 94022, USA           & Davis, CA, 95616, USA \\
			firstname.lastname@interdigital.com & hdchamain@ucdavis.edu \\
		\end{tabular}
	\end{minipage}
}

}

\maketitle

\begin{abstract}
An increasing share of captured images and videos are transmitted for storage
and remote analysis by computer vision algorithms, rather than to be viewed by humans. Contrary to traditional standard codecs with engineered tools, neural network based codecs can be trained end-to-end to optimally compress images with respect to a target rate and any given differentiable performance metric. Although it is possible to train such compression tools to achieve better rate-accuracy performance for a particular computer vision task, it could be practical and relevant to re-use the compressed bit-stream for multiple machine tasks. For this purpose, we introduce `Connectors' that are inserted between the decoder and the task algorithms to enable a direct transformation of the compressed content, which was previously optimized for a
specific task, to multiple other machine tasks. We demonstrate the effectiveness of the proposed method by achieving significant rate-accuracy
performance improvement for both image classification and object segmentation,
using the same bit-stream, originally optimized for object detection.
\end{abstract}


\section{Introduction}
Many computer vision tasks, such as object detection, tracking, image
classification, etc., rely on computationally challenging algorithms that cannot be
run on embedded devices, and require the content to be transmitted to remote
servers. Hence, the emerging research field of `Video Coding for Machines'
(VCM) aims to develop optimized compression methods for video content that is
meant to be analyzed by machines, rather than viewed by human
users~\cite{duan_video_2020}.

\begin{figure}[t]
	\setlength{\tabcolsep}{1pt}
	\footnotesize
	\centering
	\begin{tabular}{llll}
		Rate-Distortion & Rate-Detection& Rate-Distortion&Rate-Multi task\\
		\includegraphics[width=0.234\linewidth]{figures/sameBPP/bearQ1HP.png}&
		\includegraphics[width=0.234\linewidth]{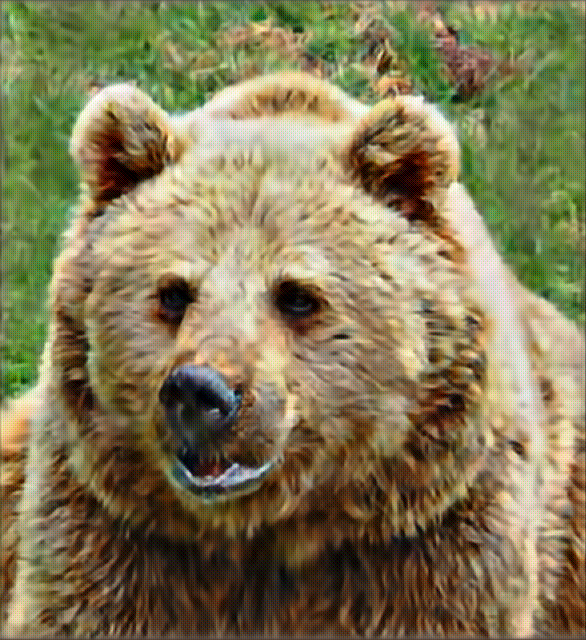}&
		\includegraphics[width=0.255\linewidth]{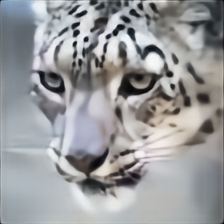}&\hspace{-1mm}
		\includegraphics[width=0.255\linewidth]{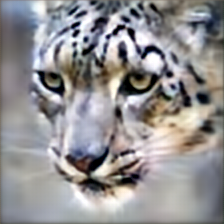}\\
		Acc mAP\@: 32.6 &34.1 (\textcolor{ForestGreen}{$+$}\textbf{4.6}\%)&Acc top-5: 66.43\%&75.18\% (\textcolor{ForestGreen}{$+$}\textbf{8.75}\%)\\
		PSNR\@: 30.49 dB &16.31 dB &26.37 dB &22.35 dB (\textcolor{red}{$-$}\textbf{4.02})
	\end{tabular}
	\vspace{1mm} 
	\caption{End-to-end optimization of an ANN-based
		codec for multiple machine tasks. Left:
		the detection accuracy is improved by 4.6\% at the same bitrate by
		optimizing the codec from~\cite{balle2018variational} for rate-detection instead of the
		conventional rate-distortion based on PSNR (`bear' image from
		COCO~\cite{lin2014microsoft}). Right: We show that it is possible to use
		a rate-`object detection' optimized codec to achieve high rate-`multi
		task accuracy' performances (classification, segmentation, etc.),
		using the proposed simple pre-trained connectors. `Tiger' from
		ImageNet-1K~\cite{russakovsky2015imagenet} is compressed using the same
		rate-`object detection' optimized codec coupled with a trained connector (81
		parameters) to achieve a 8.75\% image classification accuracy improvement on an
		off-the-shelf classifier (right image).}
	\label{fig:main}
\end{figure}

When the image/video content is analyzed by machines (e.g.  pedestrian and vehicle detection by smart cars), the codec and the considered task (we call it `primary task') algorithm can be optimized end-to-end  to achieve the optimal task accuracy at a given rate~\cite{chamain2018QuanNet}. In this paper, we study the case where the optimization of the codec for the primary task does not consider the fidelity of the decoded frames,  usually measured using Peak Signal-to-Noise Ratio (PSNR)\@. Contrary to traditional compression standards such as JPEG~\cite{jpeg}, or H.265/HEVC~\cite{hevc}, decoded images are not optimized to ensure high fidelity or viewing quality, as can be seen in Fig.~\ref{fig:main}. 
However the decoded content is still viewable~\cite{chamain2020end}, which is useful for applications that involve human supervision.
For instance, a video surveillance system may require the produced videos to be used as pieces of evidence. 

The features extracted from Convolutional Neural Networks (CNN) are the primary candidates for most visual tasks~\cite{sharif2014cnn}. However, the codecs that are optimized end-to-end for a primary task cannot be directly used on off-the-shelf secondary tasks models trained on RGB images even though they can produce features that are useful for visual machine tasks. How to adapt these features directly for visual machine tasks without piling up in complexity is still unanswered.

In this paper, we discuss how to adapt the features selected by a codec optimized for a primary task for secondary vision tasks performed on different data sets. 
In particular we propose simple, one-time trained and task-independent `Connectors' that transform primary-task-optimized-features during off-the-shelf inference for secondary visual tasks. Further we show that fine tuning the secondary task models on primary task optimized features leads to significant gain in rate-accuracy performances compared to baseline models. These observed gains result from the richness of the primary task data set, essential features learned by end-to-end optimization of the codec-primary task at a given bandwidth and the feature transformation of the task independent connector at the inference. See Fig.~\ref{fig:main}. Finally, based on the visual tasks such as object detection, classification and segmentation, we demonstrate the effectiveness of the proposed connectors with interesting visual examples. Additional visual examples and details of the experiments are given in the supplemental material.

\section{Related Works}
\label{sec:related}
In terms of optimizing image/video compression codecs for a given learning task, only a handful of works emerges from the literature.
The authors of \cite{luo2020rate} and \cite{chamain2018QuanNet} proposed to learn the quantization tables of JPEG and JPEG2000, respectively, using end-to-end training for image classification with differential codec implementations. Using a learned entropy coder, the authors of \cite{singh2020end} has shown the effectiveness  of end-to-end optimizing features for image classification. Similarly in \cite{chamain2020end}, the authors end-to-end optimize the RGB images compressed with scale hyperpriors codec~\cite{balle2018variational} for object detection on COCO data set and show significant rate-detection performance gains at lower bandwidths. In this work, we utilize the rate-detection optimized codec in \cite{chamain2020end} as the primary task optimized codec. 

In the previous works that explore for multi-task applications, we observe that tasks reusing the same resources are carefully selected to be closely related and beneficial to each other~\cite{misra2016cross}. In order to extract generic descriptors, in \cite{sharif2014cnn} authors use ImageNet data set~\cite{russakovsky2015imagenet} as it consists of over 1.2 million images and the task as fine grained classification. 

In \cite{zamir2018taskonomy}, authors assured the possibility of reusing the features among visual tasks through transfer learning without added complexity. By defining a structure connecting several visual tasks, they identify the redundancy between tasks. Furthermore the authors of \cite{luo2020rate}, discussed how the learned features generalize over different architectures for the same learning task by observing that the codec learned for the MobileNet~\cite{howard2017mobilenets} classifier achieves similar gains with a ResNet~\cite{he2016deep} classifier.

\section{Compression for multiple machine task}\label{sec:study}%

In order to find the features that are vital for common visual machine tasks such as detection, classification and segmentation etc., we propose the following strategy. First we select the important features that are non-redundant based on a single (primary) task $T$ and then transform the selected features at the inference for other (secondary) tasks $S$.

\subsection{Structure of networked codec and task algorithm}
\begin{figure}[t]
	\centering
	\includegraphics[width=\linewidth]{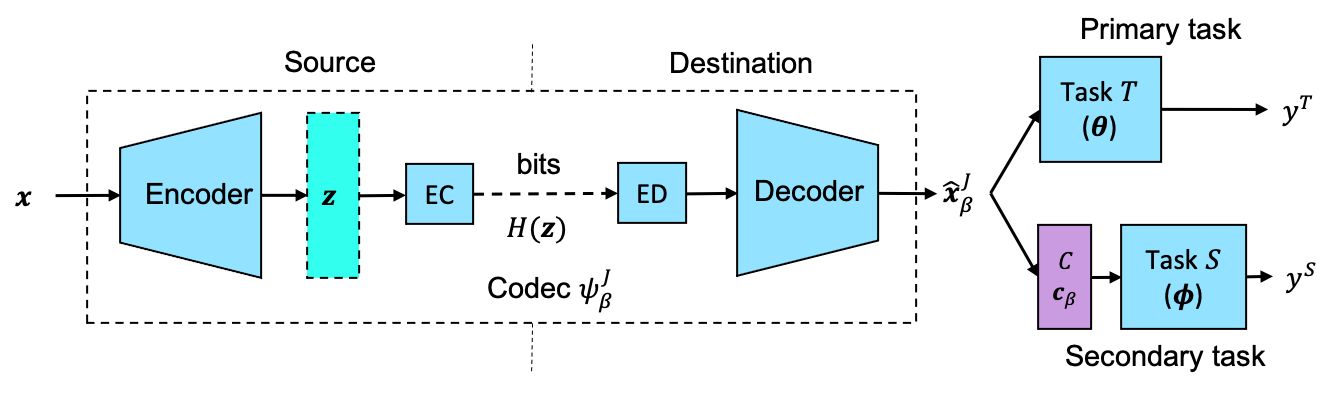}
	\caption{General structure of networked ML use case: input $\mathbf{x}$ is encoded to latent $\mathbf{z}$ and stored/transmitted over a channel as a binary stream. The decoder decodes the binary stream and reconstructs $\mathbf{\hat x}_{\beta}^{J}$ on which task $T$.is performed. Connector $C$ is applied before the inference for task $S$. }\label{fig:featureReuse}
\end{figure}

The considered networked Machine Learning (ML) inference applications observe the general structure shown in Fig.~\ref{fig:featureReuse}. It consists of a `codec' that efficiently compresses the input $\mathbf{x}$ and two `task models' that perform the primary task $T$ and the secondary task $S$ on the reconstructed images $\mathbf{\hat x}_{\beta}^{J}$. We parameterize the end-to-end (jointly) optimized codec that encodes images at a quality parameterized by $\beta$ with $\boldsymbol \psi_{\beta}^{J}$, ($\mathbf{\hat x}_{\beta}^{J} = \boldsymbol \psi_{\beta}^{J}(\mathbf{x})$) and the models for tasks $T$ and $S$ with $\boldsymbol \theta$ and $\boldsymbol\phi$. The pre-trained connector $C$ parameterized by $\boldsymbol c_{\beta}$ applied only for the secondary task $S$ at the inference.

In general, the codec consists of an Encoder Analyzer (E), an Entropy Coder (EC), an Entropy Decoder (ED) and a Decoder Synthesizer (D). See \cite{chamain2020end} for more details. During inference for task $T$, a pre-trained task model $\boldsymbol \theta$ performs the learning on $\mathbf{\hat x}_{\beta}^{J}$ and estimates the result $y^T$, ($\boldsymbol \theta(\mathbf{\hat x}_{\beta}^{J})=y^{T}$). 
In practice, the task model $\boldsymbol \phi$ is optimized for minimum task loss on the input images of best possible (reference) quality `$rq$' available during training. We denote the task model optimized for the reference quality $rq$ by $\boldsymbol \phi_{rq}$.

\subsection{Primary task selection}
Since the purpose of the primary task is to select general features for visual machine tasks, following \cite{sharif2014cnn}, we select a challenging and more general task for this. Ex: Object detection with COCO data set~\cite{lin2014microsoft}. This involves separating the image patches with objects of interest with different scales and orientations and then classifying them in to classes. With this choice of task and data set compared to the work \cite{sharif2014cnn}, we expect to extract more generic features by using the bounding box information along with class labels. 
\subsection{Task specific optimization with Joint fine tuning: J-FT}
The goal here is to fully optimize the system for the primary task. Hence,  we choose to jointly optimize the codec parameters  $\boldsymbol \psi$ and task parameters $\boldsymbol \theta$ to converge at the best rate-task accuracy performance.  Keeping the structure of the decoder which outputs RGB images enables us to directly use off-the-shelf task models without redesigning the task models to take latent features as inputs.

Assume that $\mathcal{X}_{rq}$  denotes the set containing reference quality images $\mathbf{x}\in\mathcal{X}_{rq}$ and $y_{gt}^{T} \in \mathcal{Y}^{T}$ is the ground truth for the inputs $\mathbf{x}$ for the considered task $T$. 
We express the loss function for a task $T$ with $\mathcal{L}_T(y^{T},y_{gt}^{T})$ which calculates task loss based on the prediction $y^{T}$ generated by the task model $\boldsymbol \theta$ for the input and the ground truth pair $(\mathbf{\hat x},y_{gt}^{T})$ for task $T$.

Hence, we can find the optimal codec-task model parameters combination ($\boldsymbol \psi_{\beta}^J,\boldsymbol \theta_{\beta}^J$) by minimizing the task loss and the rate~\cite{chamain2020end,chamain2018QuanNet,singh2020end}.
\begin{align}
	\label{eq:V4}
	\boldsymbol \psi_{\beta}^J, \boldsymbol \theta_{\beta}^J &= \arg \min_{\boldsymbol \psi,\boldsymbol \theta} \dfrac{1}{|\mathcal{ X}_{rq}|} \sum_{\mathbf{x} ,y_{gt}^{T} \in\mathcal{ X}_{rq}, \mathcal{Y}^{T}} \{\mathcal{L}_T(\boldsymbol \theta(\boldsymbol \psi(\mathbf{x})),y_{gt})\nonumber\\ &\qquad\qquad\qquad\qquad\qquad\qquad\qquad\quad+ \beta \mathcal{L}_R(\mathbf{z})\}
\end{align}

$\mathcal{L}_R(\mathbf{z})$ denotes the entropy $H(\mathbf{z})$ of the latent representation $\mathbf{z}$. See Fig~\ref{fig:featureReuse}. The control parameter $\beta$ regulates the trade-off between the accuracy of task $T$  and the bandwidth/rate of the input to the task model.

\subsection{Introducing Connectors}
\label{sec:reuse-connectors}
Even though the optimally selected features for a given bandwidth with the primary task are important for a secondary task $S$, the scale of the features are not fine tuned for the pre-trained models of task $S$ that expects images of reference quality $\mathbf{\hat x}_{rq}$ as inputs. Hence, the end-to-end optimized inputs $\mathbf{\hat x}_\beta^J$ perform poorly in terms of rate-accuracy performance for task $S$ on a pre-trained off-the-shelf task model $\boldsymbol \phi_{rq}$ trained with rate-distortion optimized inputs. 

In order to de-specialize the optimized images $\mathbf{\hat x}_\beta^J$, we add a connector $C$ parameterized by $\boldsymbol c$ in between the codec output $\mathbf{\hat x}_\beta^J$ and the task model $\boldsymbol \phi_{rq}$ to obtain $\mathbf{\hat x}^C_\beta$ which is the input to the model $\boldsymbol \phi_{rq}$. 
In the context of re-usability of features that are originally optimized for a primary task, on off-the-shelf secondary task models, the codec $\boldsymbol \psi_\beta^J$ and the pre-trained model $\boldsymbol{\phi}_{rq}$ are fixed, hence beyond out control. We explore the possibility of a simple fixed/trainable connector that can be stored at the end devices that perform inference for multiple tasks. 

For trainable $C$ we propose to find the optimal parameter setting $\boldsymbol c_{\beta}^{J}$ that minimizes the visual distortion $\mathcal{L}_D$ between the connector produced image $\mathbf{\hat x}_\beta^C$ and its rate-distortion optimized version $\mathbf{ x}_{rq}^D$. Therefore we find $\boldsymbol c_{\beta}^{J}$ s.t.
\begin{align}
	\label{eq:V5}
	\boldsymbol c_{\beta}^{J} = \arg \min_{\boldsymbol c}  \dfrac{1}{|\mathcal{X}_{rq}|}\sum_{\mathbf{ x}\in \mathcal{X}_{rq}}  \mathcal{L}_D(\boldsymbol c(\boldsymbol \psi_{\beta}^J(\mathbf{x})),\mathbf{ x}).
\end{align}

For each rate-accuracy control parameter $\beta$, the trained connector $\boldsymbol c_{\beta}^{J}$ can be stored at the decoder and used for images that were compressed with the codec $\boldsymbol \psi_\beta^J$ during inference of any secondary task according to the following flow for the input, ground truth pair ($\mathbf{x}$, $y_{gt}^{S}\in \mathcal{Y}^{S}$). Here $y^{S}$ is the estimated label for task $S$.
\begin{equation}
	\mathbf{x} \xrightarrow[\boldsymbol \psi_\beta^J]{\text{E-EC-ED-D}} \mathbf{\hat x}_\beta^J\xrightarrow[\boldsymbol c_{\beta}^{J}]{C} \mathbf{\hat x}^C_\beta\xrightarrow[\boldsymbol \phi_{rq}]{S}y
\end{equation}
 Note that the proposed optimization of $\boldsymbol c$ is independent of the secondary task $S$. Therefore training of the connector given in Eq.~\eqref{eq:V5} is only performed once and zero knowledge of the secondary task is required.
\subsection{Fine-tuning the task model with end-to-end optimized images: J-FT T-FT}
Given that we have the access to the secondary task model and training set, for better accuracy at task $S$, we can fine tune the task model $\boldsymbol \phi$ on images compressed and reconstructed using the end-to-end jointly optimized coded $\boldsymbol \psi_\beta^J$.

The fine-tuned task model $\boldsymbol \phi_{\beta}$ is obtained by minimizing the task loss $\mathcal{L}_S$ as the following at each rate specified by $\beta$.
\begin{align}
	\boldsymbol \phi_{\beta} = \arg \min_{\boldsymbol \phi}  \dfrac{1}{|\mathcal{X}_{rq}|}\sum_{\mathbf{ x} ,y_{gt}^{S} \in\mathcal{ X}_{rq}, \mathcal{Y}^S} \mathcal{L}_S(\boldsymbol \phi(\boldsymbol \psi_{\beta}^J(\mathbf{x})),y_{gt}^{S})
\end{align}
\section{Experiments and Results}
As the primary task $T$, we used the well researched object detection on COCO-2017~\cite{lin2014microsoft} data set.
COCO-2017 dataset for object detection contains over 180k training and 5k validation RGB images of average size over 256$\times$256 with annotated bounding boxes belonging to 80 classes. We used the standard primary challenge metric `box mAP' 
to evaluate the detection accuracy of the val set and the model Faster-RCNN~\cite{ren2015faster} with ResNet-50 backbone following \cite{chamain2020end} to reproduce their results shown as J-FT in \cite{chamain2020end}.

Similarly as the image codec for this paper, we used Scale hyperpriors model \textit{bmshj2018-hyperprior}, optimized on MSE~\cite{balle2018variational} which is an auto encoder-decoder based differentiable codec featuring a learned entropy coder.

\subsection{End-to-end codec optimization for the primary task}
Following Eq.~\eqref{eq:V4}, we fine tuned the codec and the detector for 6 epochs with the same optimizers and learning rates used in \cite{chamain2020end} and regenerated the results (shown in Fig.~3 of \cite{chamain2020end}). Note that for joint end-to-end training, the rate-detection performance gain is significant compared to codec only (C-FT) and task only (T-FT) fine tunning at lower bit rates.

\subsection{{Application of Connectors}}
As the secondary tasks for inference, we selected image classification ($S_1$) on ImageNet-1k~\cite{russakovsky2015imagenet}  and semantic segmentation ($S_1$) on Cityscapes~\cite{cordts2016cityscapes}. ImageNet-1K consists of RGB images of size over 256$\times$256 belonging to 1000 classes, each of them containing about 1300 training and 50 validation samples. We evaluated the performance of the classification task using top-1 and top-5 accuracy as metrics. Cityscapes is a video dataset recorded in streets from 50 cities captured in 5000 frames split into 2975 training, 500 validation and 1525 test samples. For our experiments we used the left frames of training and validation sets with fine annotations belonging to 19 classes. The task of per-pixel semantic labeling is to label each pixel into one of 19 classes measured by the metric IoU~\cite{cordts2016cityscapes}.

We explore the performance of 3 simple blocks as candidates for the connecor: {\bf{1) C-Avgpool}}: 2D average pooling block of size 3$\times$3 (ablation) {\bf{2) C-DepthConv}}: 2D depth-wise convolution block 3$\times$3 (27 parameters) {\bf{3) C-Conv}}: 2D convolution block of size 3$\times$3 (81 parameters). To train the connectors C-deptConv and C-Conv, we used COCO-2017 validation set as ground truths. According to Eq.~\eqref{eq:V5}, we minimized the MSE (as $\mathcal{L}_D$) between the ground truths and the connector output images using the Adam optimizer with an initial learning rate of 0.1 reduced by a factor of 0.1 after each 10 epochs for a total of 30 epochs.
\begin{figure*}
	\centering
	\subfloat[ImageNet-1K]{\includegraphics[trim=10 10 40 20,clip,width=0.480\textwidth]{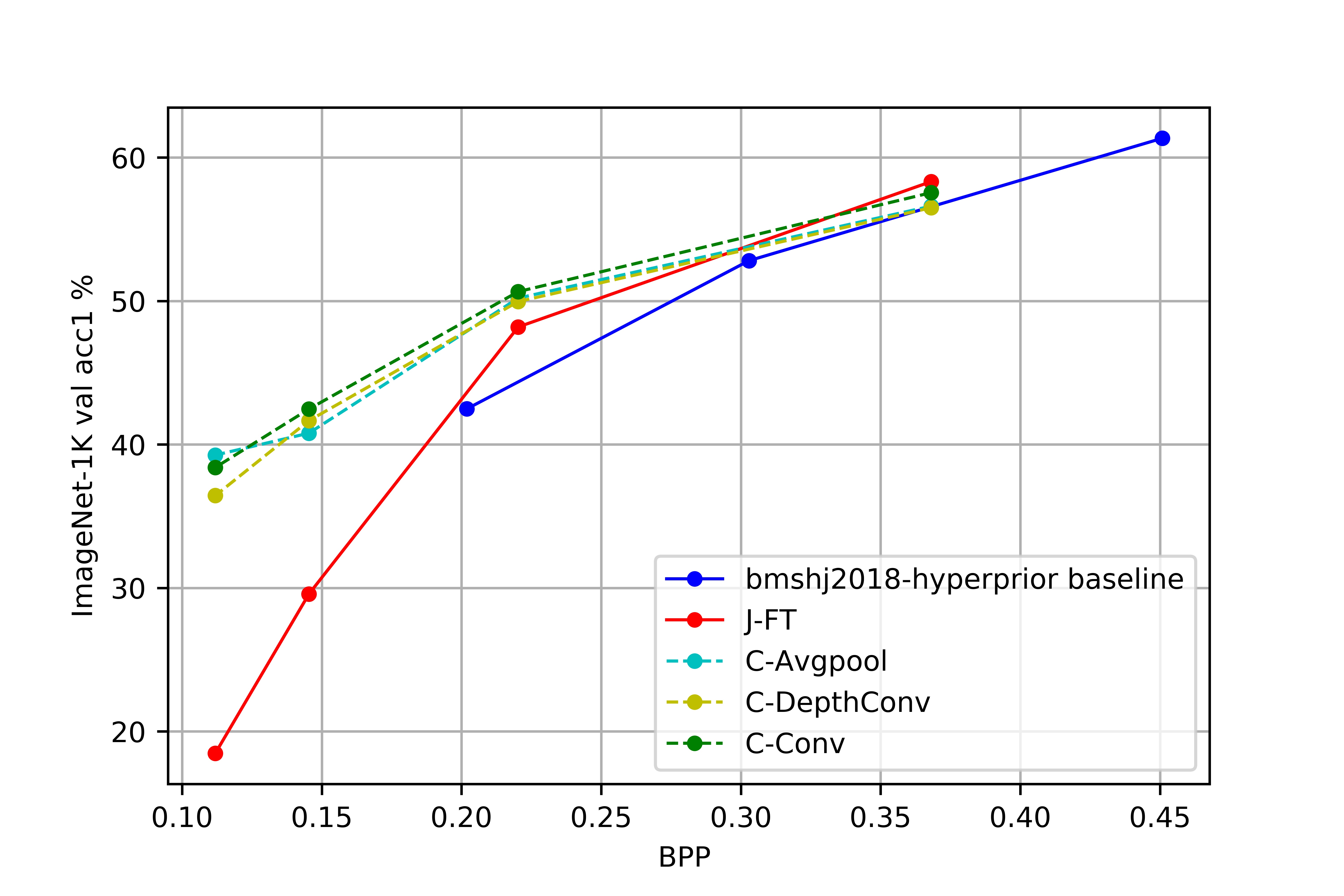}}\hfill
	\subfloat[Cityscapes]{\includegraphics[trim=10 10 40 20,clip,width=0.480\textwidth]{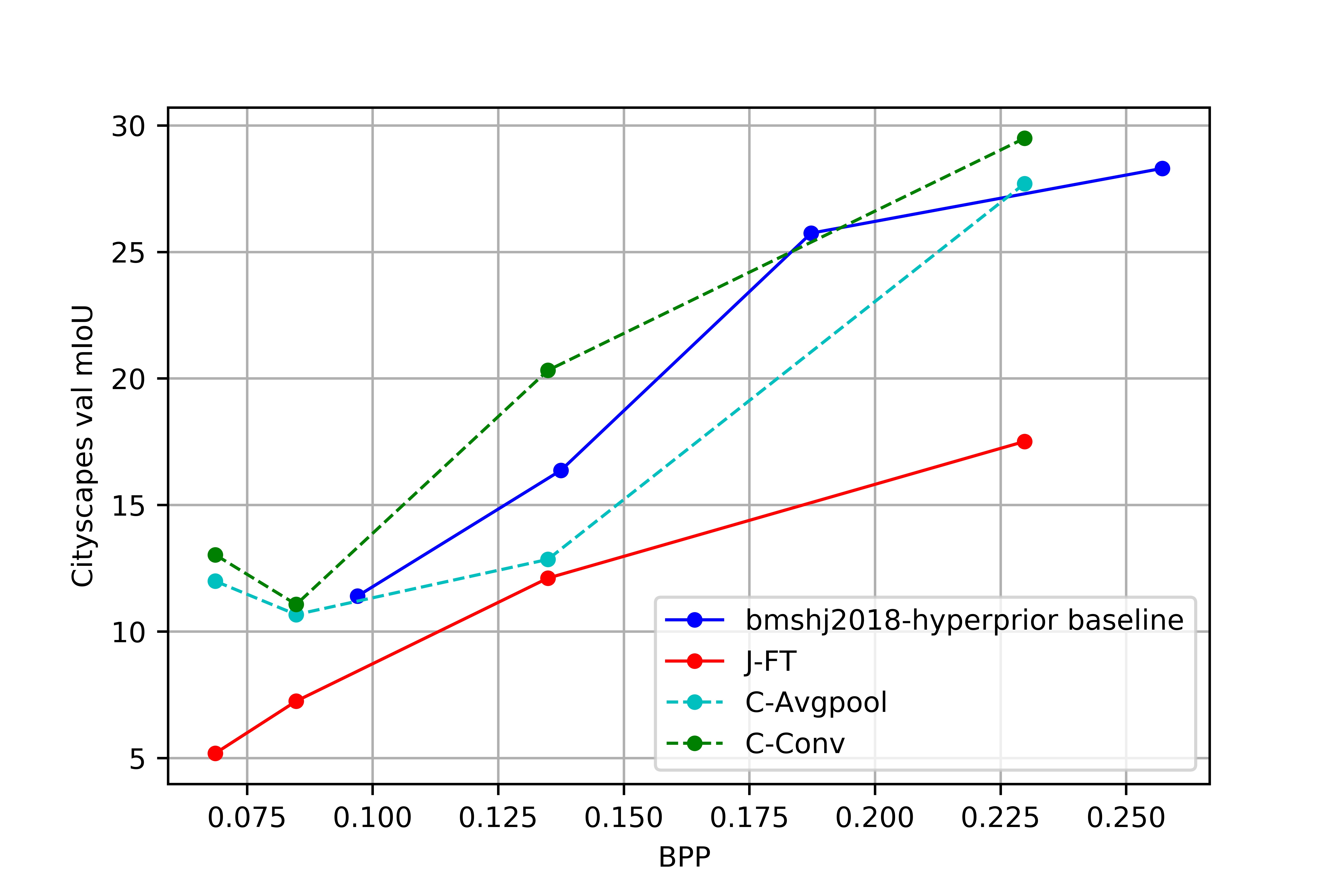}}
	\vspace{1mm}
	\caption{Inference results on the validation sets with connectors. baseline: distortion optimized images, J-FT: images compressed and decompressed by the codec optimized for detection, C-Avgpool,  C-DepthConv, C-Conv: J-FT images after each connector}\label{fig:imagenetCOCO}
\end{figure*}

Fig.~\ref{fig:imagenetCOCO}(a) shows the top-1 accuracy on the ImageNet-1K validation set with off-the-shelf ResNet-50 classifier. Interestingly, the features optimized for COCO-detection achieve better classification accuracy compared to the images generated by rate-distortion optimized codecs on off-the-shelf ResNet trained with rate-distortion optimized images. Image classification being part of COCO-detection task can be a reason for this behavior. Note that all 3 connectors achieve similar gains for ImageNet-1K inference which is significantly higher compared to the extrapolated inference on rate-distortion optimized images (baseline) at lower data rates.

Similarly Fig.~\ref{fig:imagenetCOCO}(b) shows the results for semantic segmentation on Cityscapes data set. For inference, we used a Deeplabv3+~\cite{chen2018encoder} with ResNet-50 backbone trained from scratch on the training set of the `left 8 bit' original images with fine annotations as the ground truths. Using the trained model, we repeated the similar inference steps on Cityscapes val set with the connectors trained with COCO val set. 

The same pre-trained connectors are used on ImageNet-1K and Cityscapes. Contrary to ImageNet-1K results, C-Conv connector shows significant gains for the segmentation task on Cityscapes. Overall, the  pre-trained connector enables the reuse of decoded images (optimized for a primary task) for secondary tasks even with better rate-accuracy performance.

\subsection{Effect of connectors on visualization}\label{sec:Discussion}%
\begin{figure}[t]
	\setlength{\tabcolsep}{1pt}
	\centering
	\footnotesize
	\begin{tabular}{rccc}
		Original & J-FT\cite{chamain2020end}&C-Avgpool& C-Conv\\	
		\includegraphics[width=0.24\linewidth]{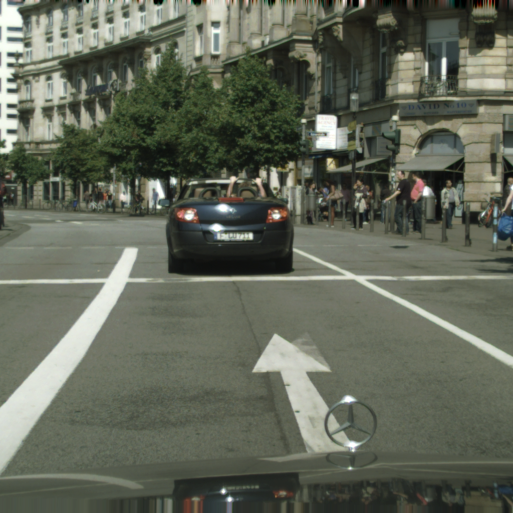}&
		\includegraphics[width=0.24\linewidth]{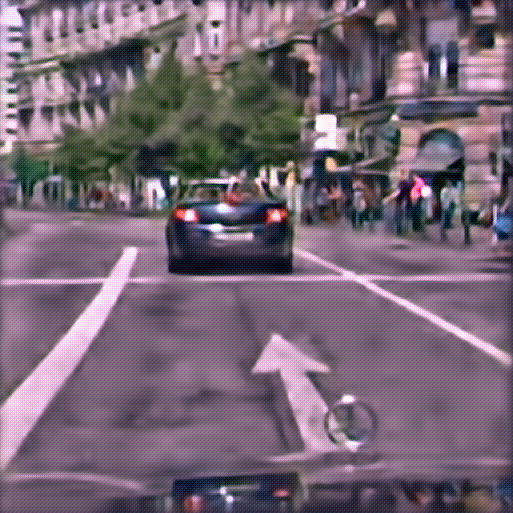}&
		\includegraphics[width=0.24\linewidth]{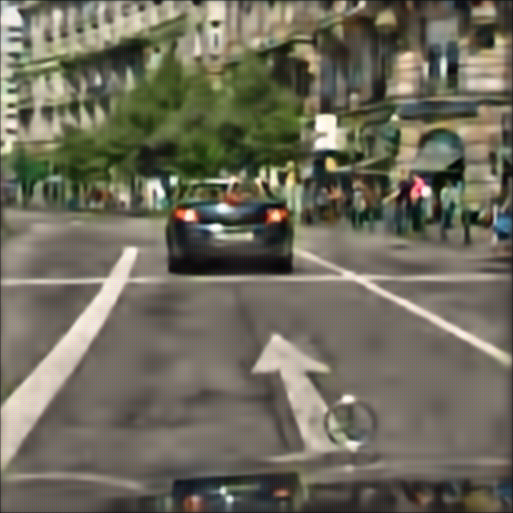}&
		\includegraphics[width=0.24\linewidth]{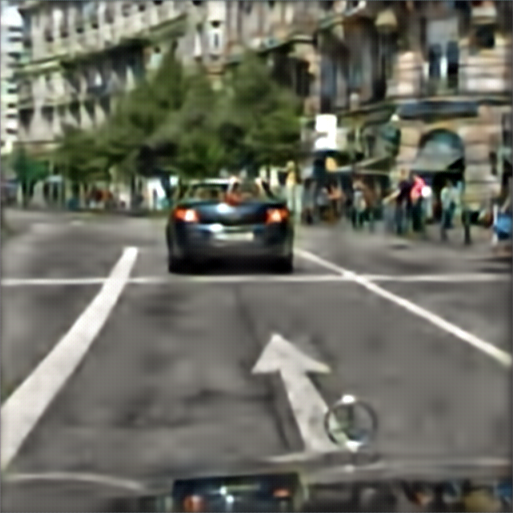}
		\\
		PSNR&17.93&21.34&21.86\\
		BPP&0.0687&0.0687&0.0687\\
		&
		\includegraphics[width=0.24\linewidth]{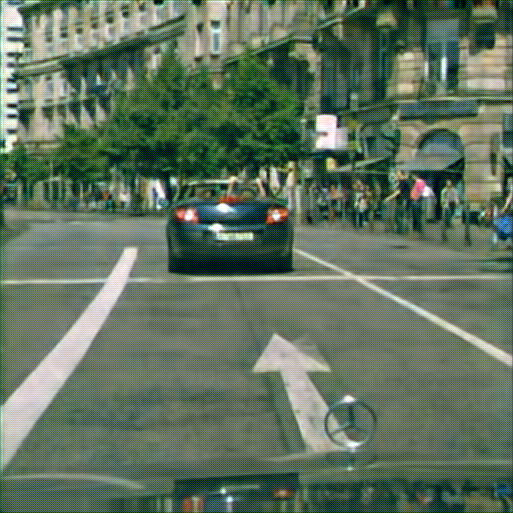}&
		\includegraphics[width=0.24\linewidth]{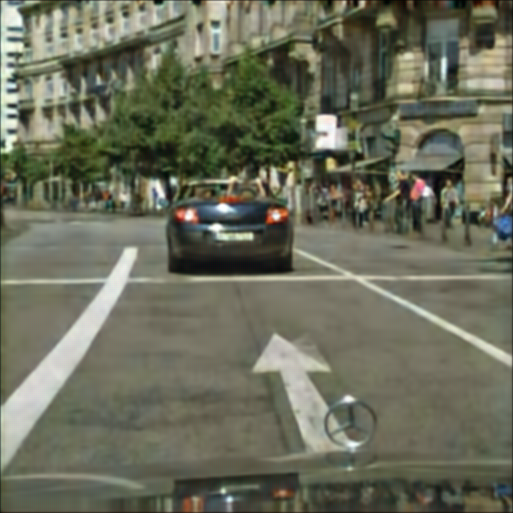}&
		\includegraphics[width=0.24\linewidth]{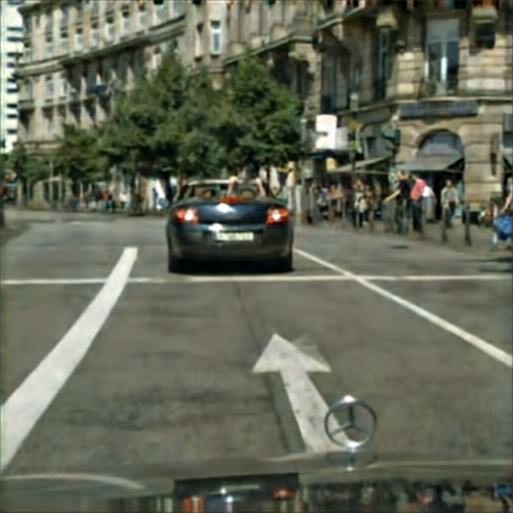}
		\\
		PSNR&20.73&25.03&27.36\\
		BPP&0.2298&0.2298&0.2298\\
	\end{tabular}
\vspace{1mm}
\caption{Comparison of the visual quality of an image from Cityscapes data set. Columns J-FT~\cite{chamain2020end}: Rate-detection optimized image. C-Avgpool, C-Conv: output of each connector.}\label{fig:citycar}
\end{figure}

Since Eq.~\eqref{eq:V4} does not include any criterion related to the visual quality
of decoded images, or distortion constraints with respect to the original, the codec optimized
for object detection (J-FT~\cite{chamain2020end}) produces low fidelity images. Fig.~\ref{fig:citycar} shows the reconstructed images from the output of
C-Avgpool and C-Conv. It is clear that J-FT produces images that include high frequency features, which
help detecting objects.  The proposed connectors generally smoothen the details of decoded images, specific to the primary task,  making them more generic for any secondary task or viewing. 
Since the proposed training of the connector aim to minimize the MSE between rate-detection optimized and rate-distortion optimized images, training of the connector once per given rate (BPP) is sufficient.

In both cases, connectors output higher quality compared to J-FT~\cite{chamain2020end} both visually and
quantitatively (PSNR). Further examples and details including PSNR and MS-SSIM curves on the COCO dataset, are provided in supplement. 

\subsection{Fine-tuning the classifier for better performance}
Being able to train/fine-tune the secondary task model, we observe that the latent/feature maps optimized for a primary task (detection) achieves significant accuracy gains for the secondary tasks from fine-tuning as well compared to fine-tuning on rate-distortion optimized images.

\begin{figure*}
	\centering
	\subfloat[ImageNet-1K]{\includegraphics[trim=10 10 40 20,clip,width=0.480\textwidth]{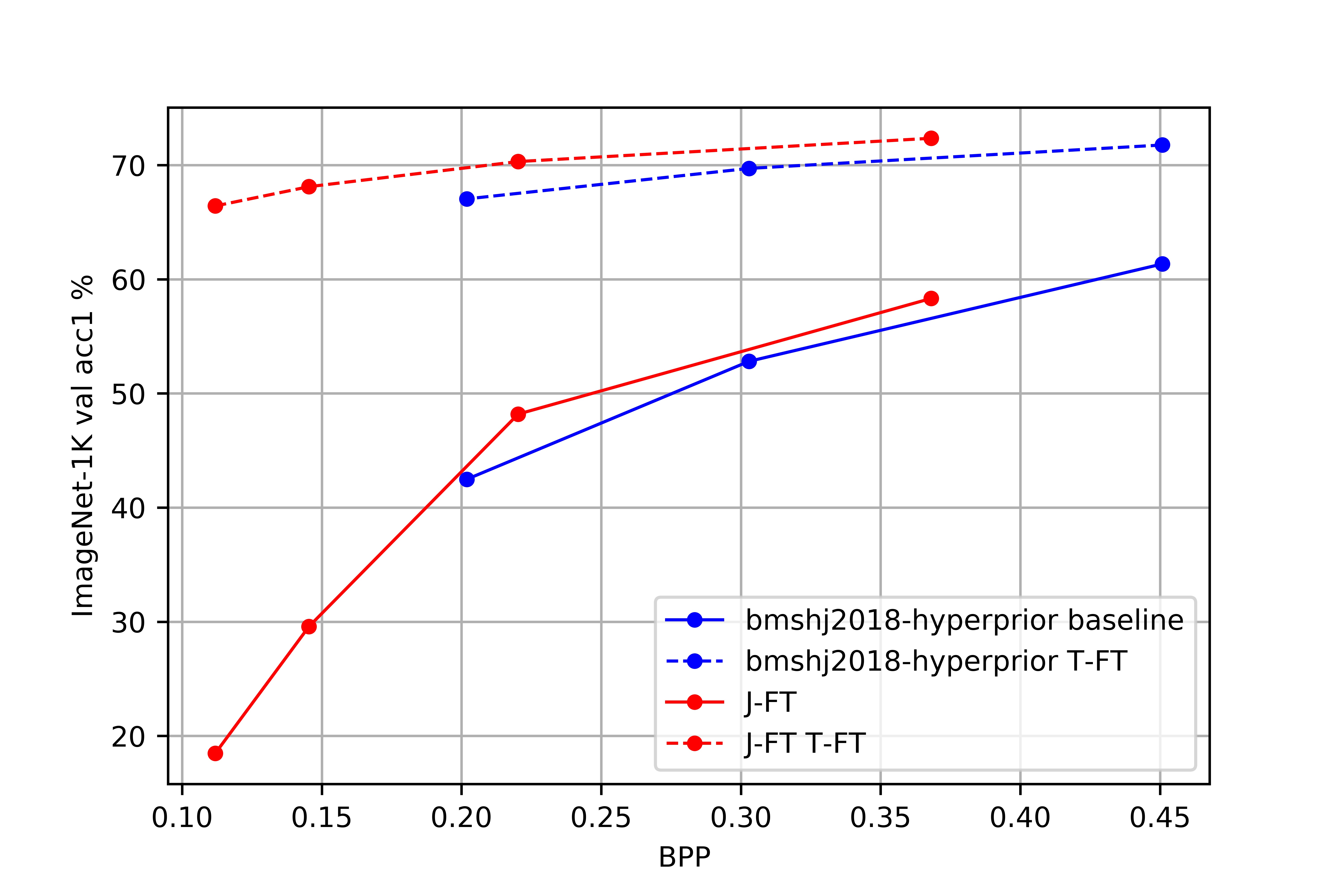}}\hfill
	\subfloat[Citysapes]{\includegraphics[trim=10 10 40 20,clip,width=0.480\textwidth]{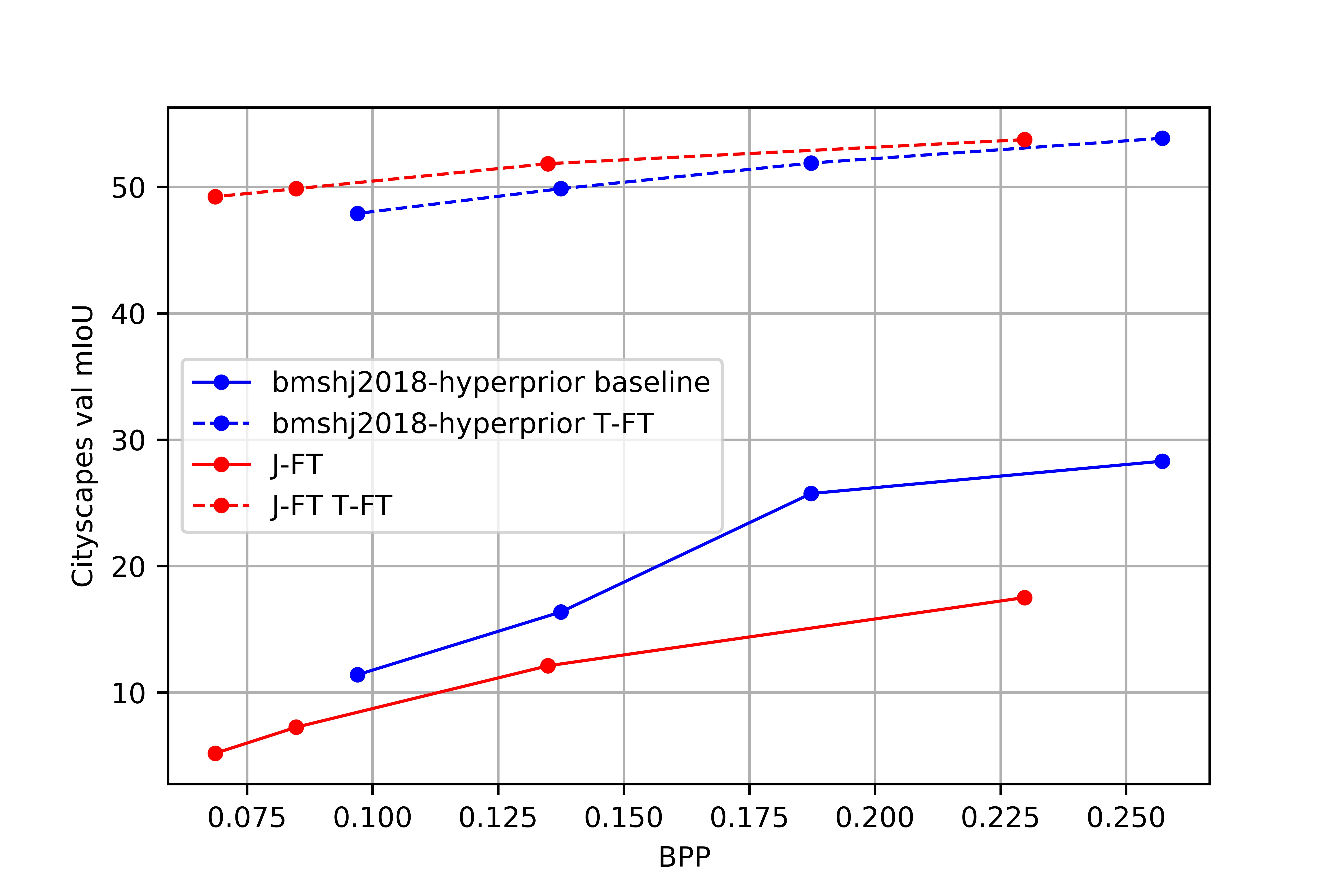}}
	\vspace{1mm}
	\caption{Fine-tuning the classifier. T-FT: fine-tuned task on distortion optimized decoded images. J-FT: pre-trained task on rate-detection optimized images. J-FT T-FT: fine-tuned task on rate-detection optimized images.}\label{fig:imagenetCOCO2}
\end{figure*}
To demonstrate the validity of this claim, we fine-tuned the secondary task models with rate-detection optimized images separately at each rate point (J-FT T-FT). As a baseline to compare with, we fine-tuned the secondary task models (T-FT) on images compressed with rate-distortion optimized \textit{bmshj2018-hyperprior}. Fig.~\ref{fig:imagenetCOCO2}(a)~ and ~\ref{fig:imagenetCOCO2}(b) compares the accuracy gains for ImageNet-1K and Cityscapes respectively.

We fine-tuned the classifier using an SGD optimizer minimizing the classification cross entropy at a learning rate of 0.001 and a momentum of 0.9 for 5 epochs for both \textit{bmshj2018-hyperprior} T-FT and J-FT T-FT settings. For \textit{Cityscapes} segmentation, we used an SGD optimizer for 50 epochs, with a momentum of 0.9 and a polynomial learning rate decay strategy initialized at 0.01 with a decay of 5e-4.

For instance, the proposed method achieves 70\% top-1 classification accuracy at 0.21 BPP which corresponds to  55\% bandwidth saving compared to rate-distortion optimized \textit{bmshj2018-hyperprior} T-FT. The J-FT T-FT setting records 3\% classification accuracy improvement at 0.2 BPP

\section{Conclusion}\label{sec:conclusion}
In this paper, we introduced simple pre-trained connectors, adapting the images decoded from a compressed bit-stream originally optimized for object detection to other secondary vision tasks, such as image
classification and segmentation. Based on experiments on ImageNet-1K and Cityscapes data sets, we showed that the proposed connectors achieve significantly better rate-accuracy performance compared to the conventional codecs optimized for rate-distortion. In future work, we plan to explore the applicability of those connectors to video compression for multiple machine tasks.
\bibliographystyle{IEEEtran}
\bibliography{main_bib}
\newpage

\begin{abstract}
	In this document, we provide additional details on the experiments we carried out to help reproduce the results presented in the paper. Furthermore, additional sample images reconstructed from COCO, ImageNet-1K and Cityscapes data sets are shown to demonstrate the effectiveness of the proposed connectors.
\end{abstract}

\begin{IEEEkeywords}
	Image compression, multiple machine tasks
\end{IEEEkeywords}

\section{Rate-Detection Baseline }

We experimented with Scale hyperprior image compression code originally optimized for Mean Square Error (MSE).

\subsection{Scale hyperprior}
\begin{figure}[ht]
	\centering
	\includegraphics[width=0.9\linewidth]{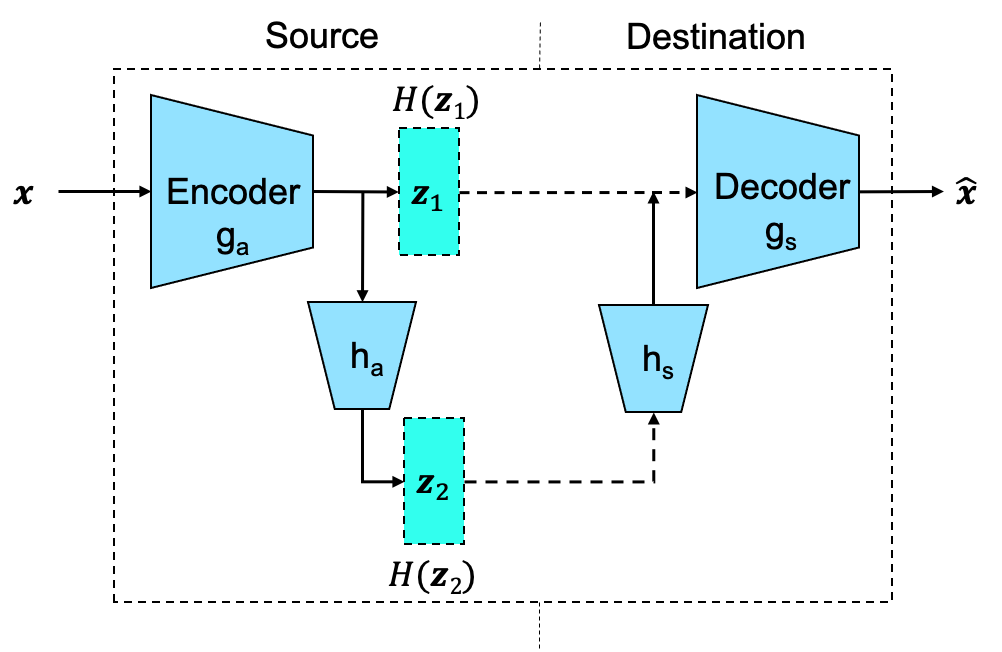}
	\caption{Structure of Scale hyperprior codec}
	\label{fig:ScaleHP}
\end{figure}

We used the implementation of \textit{bmshj2018-hyperprior} method from \cite{balle2018variational}, of which a pytorch version is available at \url{https://github.com/InterDigitalInc/CompressAI} with MSE-optimized pre-trained models.
Fig.~\ref{fig:ScaleHP} shows the basic building blocks of the scale hyperprior codec. For any input image $\mathbf{x}$, the Encoder analysis $g_a$ produces a quantized tensor $\mathbf{z}_1$, containing the mean values of the coefficients of the latent, and their standard deviations (known as scale hyperpriors) $\mathbf{z}_2$. The tensors $\mathbf{z}_1$ and $\mathbf{z}_2$ are encoded using arithmetic coding and the bit stream is transmitted or stored. At the destination, $g_s$ synthesizes the reconstructed image $\mathbf{\hat x}$ from the arithmetically decoded $\mathbf{z}_1$ and $\mathbf{z}_2$. The total Rate of the bit stream for each image is the sum of the Entropies $H(\mathbf{z}_1)$ and $H(\mathbf{z}_2)$.

The blocks $g_a$, $h_a$, $g_s$ and $h_s$ consist of 2d-convolutional layers and generalized divisive normalization (GDN/IGDN) layers. For the codec optimization with a fixed decoder setting (J-FT-FD), we only updated the parameters in $g_a$ and $h_a$during training.

\subsection{Faster-RCNN model}
As the detection model, we used the off-the shelf Faster-RCNN with ResNet-50~\cite{he2016deep} backbone available at \url{https://github.com/open-mmlab/mmdetection}. In order to achieve better detection performance, COCO-2017 validation images are resized (scaled-up) to the resolution $800 \times 1200$ at the pre-processing stage before feeding to the detector.\\

\section{Visual artifacts on J-FT optimized images}
\begin{figure*}[!htbp]
	\setlength{\tabcolsep}{1pt}
	\centering
	\begin{tabular}{ccccc}
		$\psi_q^D$&
		\includegraphics[width=0.21\linewidth]{figures/featureReuse/bearQ3HP.png}&
		\includegraphics[width=0.21\linewidth]{figures/featureReuse/bearQ3HP.png}&
		\includegraphics[width=0.21\linewidth]{figures/featureReuse/bearQ2HP.png}&
		\includegraphics[width=0.21\linewidth]{figures/featureReuse/bearQ1HP.png}\\
		&$q=4$&$q=3$&$q=2$&$q=1$\\	
		BPP&0.2875&0.2054&0.1461&0.1024\\
		$\psi_{\beta}^J$&
		\includegraphics[width=0.21\linewidth]{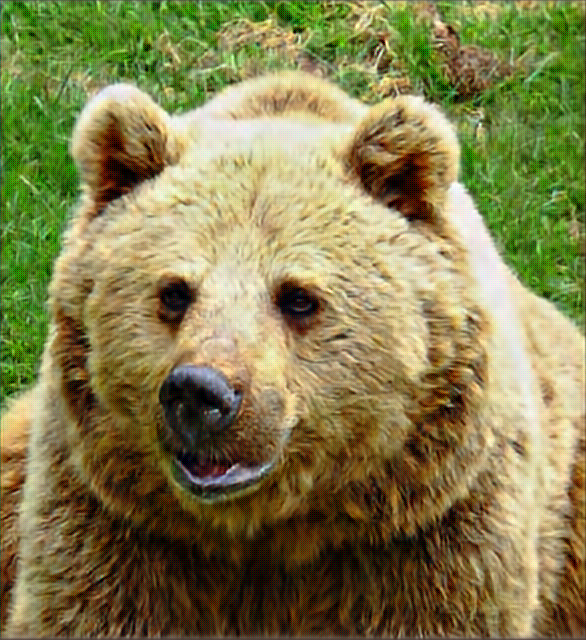}&
		\includegraphics[width=0.21\linewidth]{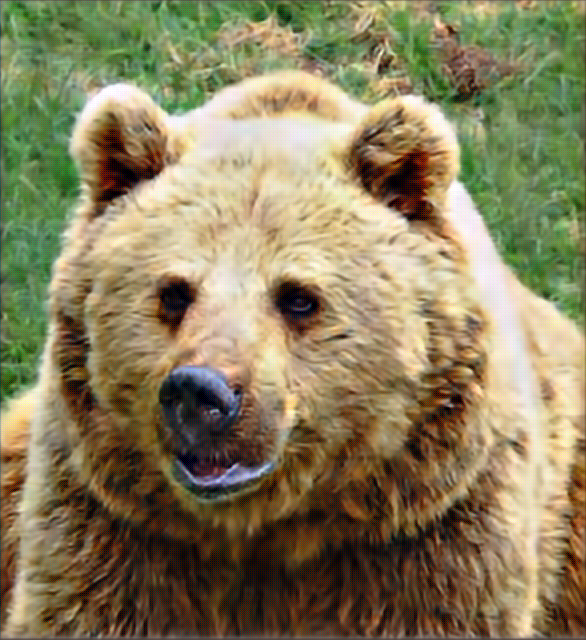}&
		\includegraphics[width=0.21\linewidth]{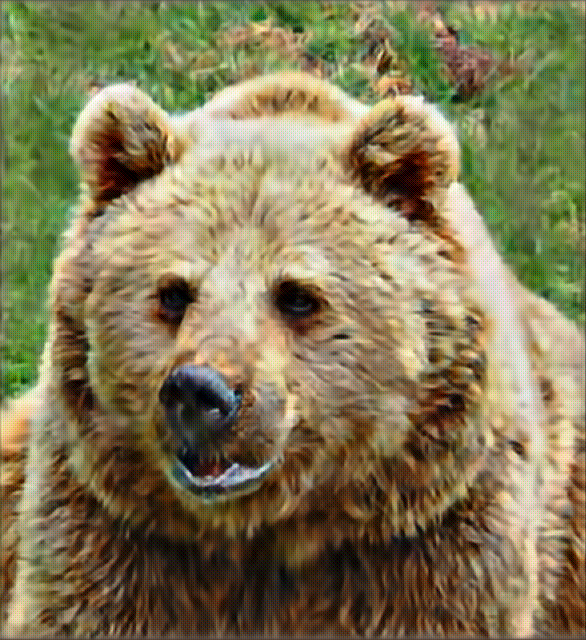}&
		\includegraphics[width=0.21\linewidth]{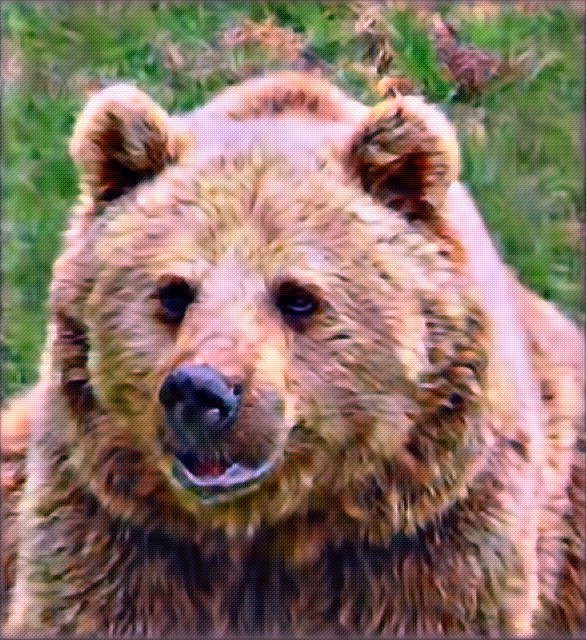}\\
		&$\beta=0.1$&$\beta=0.3186$&$\beta=0.6675$&$\beta=1.0$\\
		BPP&0.2472&0.1453&0.0943&0.0751\\
		$\psi_q^{J\rm{-FT-FD}}$&
		\includegraphics[width=0.21\linewidth]{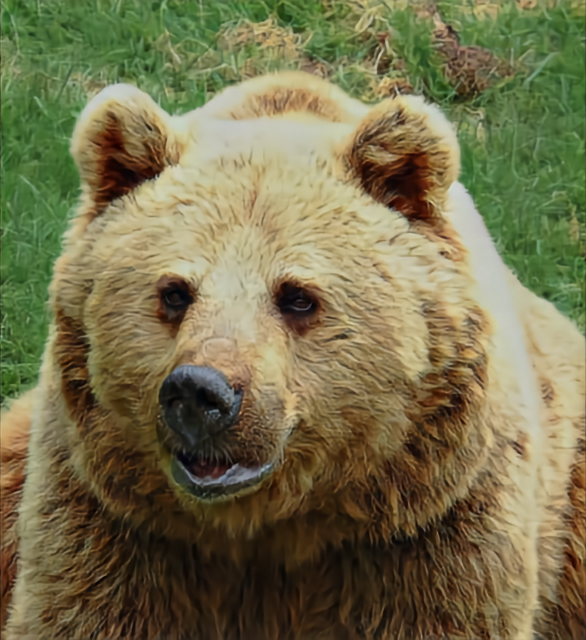}&
		\includegraphics[width=0.21\linewidth]{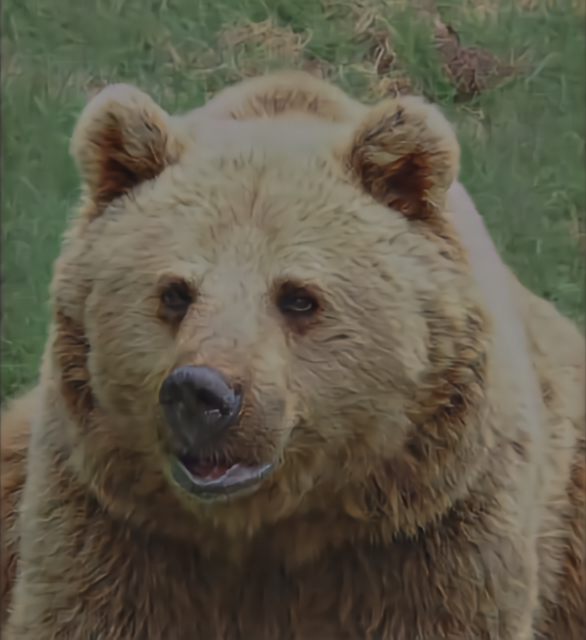}&
		\includegraphics[width=0.21\linewidth]{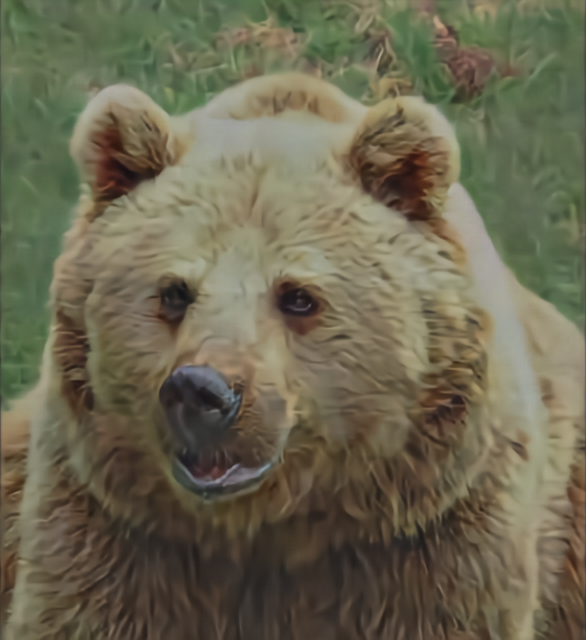}&
		\includegraphics[width=0.21\linewidth]{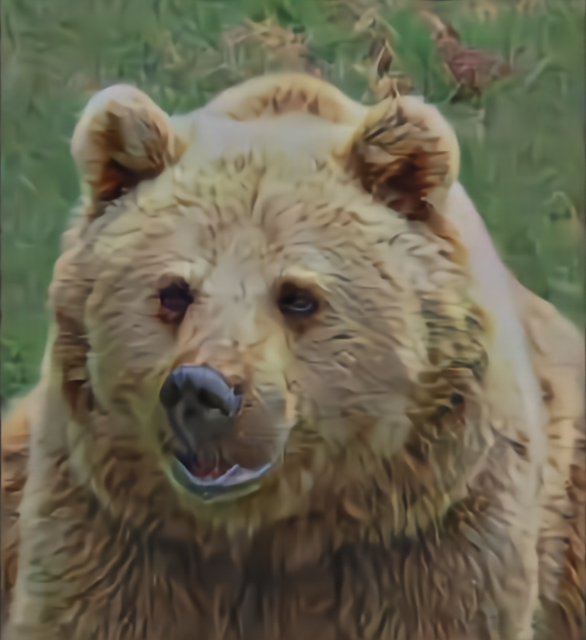}\\
		&$\beta=0.1$&$\beta=0.3186$&$\beta=0.6675$&$\beta=1.0$\\
		BPP&0.2818&0.1662&0.1069&0.0813\\
	\end{tabular}
	\caption{Visible artifacts on end-to-end optimized images. }
	\label{fig:jointrecon}
\end{figure*}

In Fig.~\ref{fig:jointrecon} we compare the visual quality of the images generated with the rate-distortion optimized Scale-hyperprior codec $\psi_q^D$ as described in \cite{chamain2020end} and the end-to-end rate-detection optimized Scale-hyperprior codec $\psi_{\beta}^J$. Images decoded from the end-to-end method optimized for object detection clearly show visible artifacts, which highlights the features used for detection. These artifacts result from the optimization based on compression rate and detection accuracy, using a loss function that does not account for the (visual) distortion.

In addition to visible artifacts, the images generated with the end-to-end optimized codec with a fixed decoder show color distortions, since the decoder has been optimized for rate-distortion.
\section{Feature reuse with connectors}
\begin{figure*}[!htbp]
	\centering
	\subfloat[top-1]{\includegraphics[width=0.5\linewidth]{figures/imagenetacc1V5_4.jpg}} \subfloat[top-5]{\includegraphics[width=0.5\linewidth]{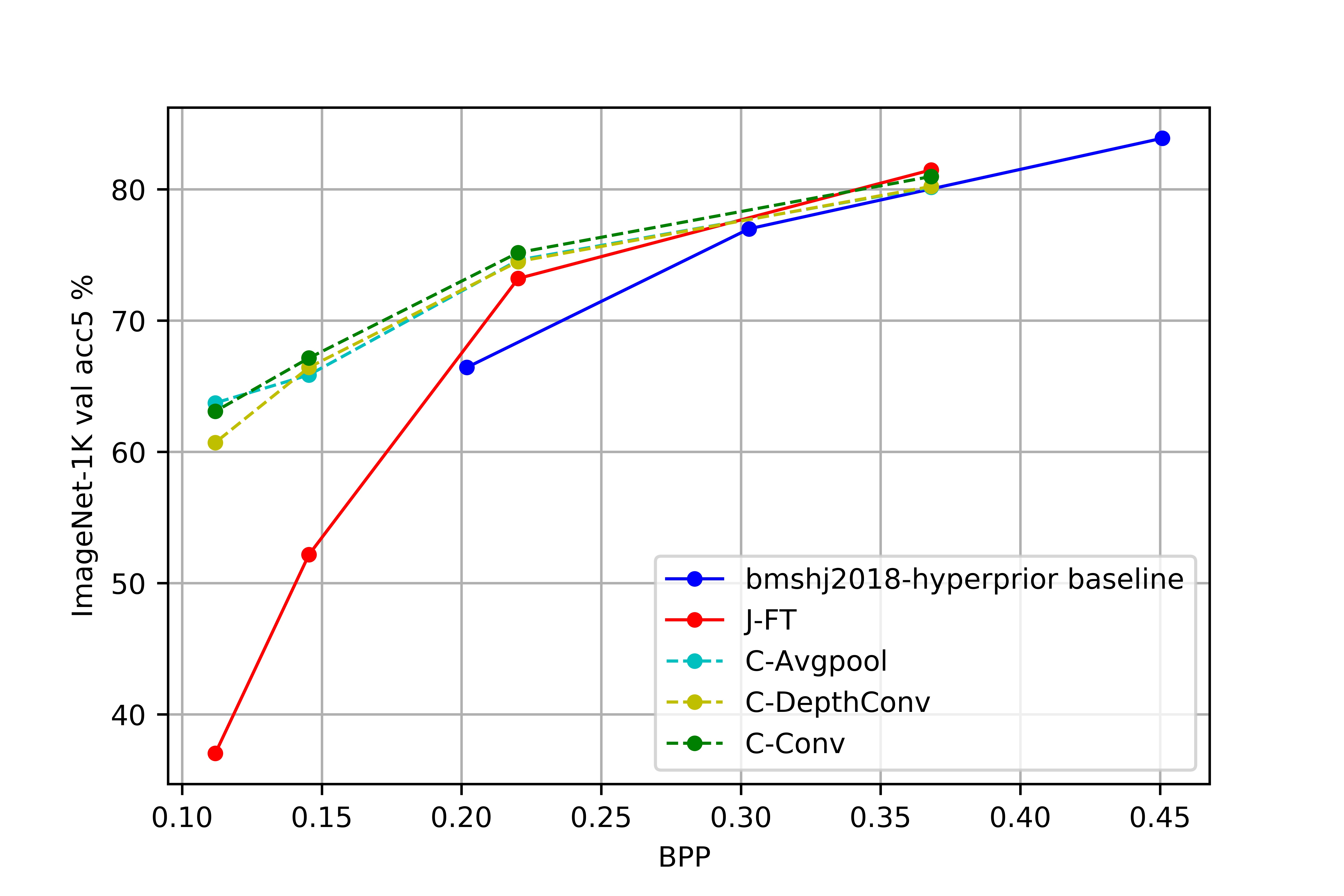}}
	\caption{Inference on ImageNet-1K. baseline: distortion optimized images, J-FT : images compressed and decompressed with codec optimized for detection (J-FT), C-Avgpool: J-FT images after average pooling, C-DepthConv: J-FT images after 2D depth-wise conv. connector, C-Conv: J-FT w/o C images after 2D conv.}
	\label{fig:supimagenet1}%
\end{figure*}

In this section we present some additional results that help benchmark the performance of the proposed connectors. Fig.~\ref{fig:supimagenet1} illustrates the  top-1 and top-5 ImageNet-1K~\cite{russakovsky2015imagenet} classification accuracy on the validation set. Note that both metrics top-1 and top-5 show similar improvements at the rates $\leq$ 0.225 BPP.

\begin{figure*}[!htbp]
	\centering
	\subfloat[top-1]{\includegraphics[width=0.5\linewidth]{figures/imagenetacc1V6_3.jpg}} \subfloat[top-5]{\includegraphics[width=0.5\linewidth]{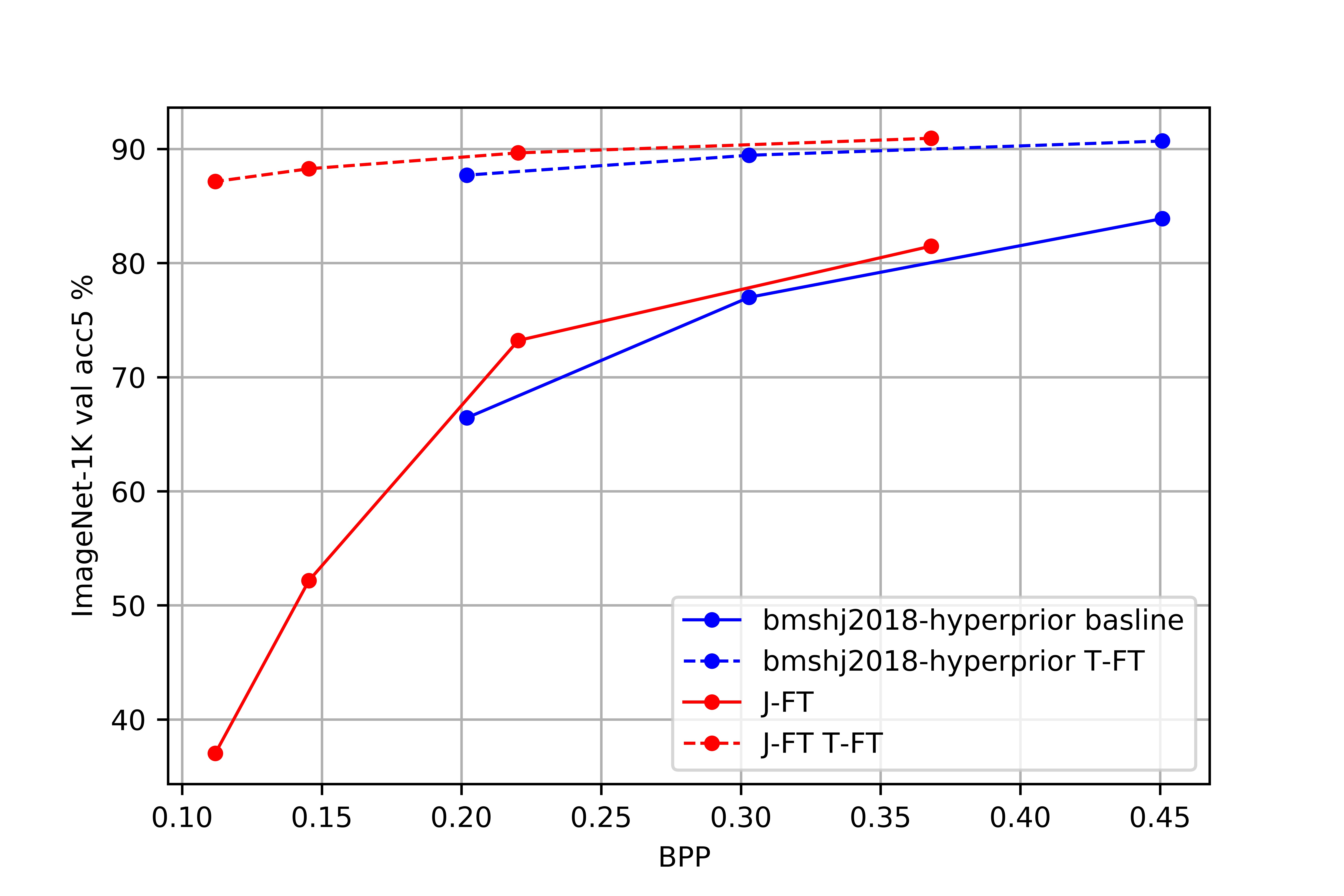}}
	\caption{Fine-tuning the classifier on rate-detection optimized ImageNet-1K. baseline: inference  T-FT: fine-tuning on distortion optimized images. J-FT : inference on rate-detection optimized images J-FT T-FT: fine-tuning on rate-detection optimized images.}
	\label{fig:supimagenet2}%
\end{figure*}

Similarly, Fig.~\ref{fig:supimagenet2} gives the top-1 and top-5 ImageNet-1K classification accuracy on the validation set when fine tuning a ResNet-50 classifier. We observe that fine-tuning the classifier with the images that are end-to-end optimized for detection facilitates the classification, compared to rate-distortion optimized images.

\begin{figure*}[!htbp]
	\centering
	\subfloat[PSNR]{\includegraphics[width=0.5\linewidth]{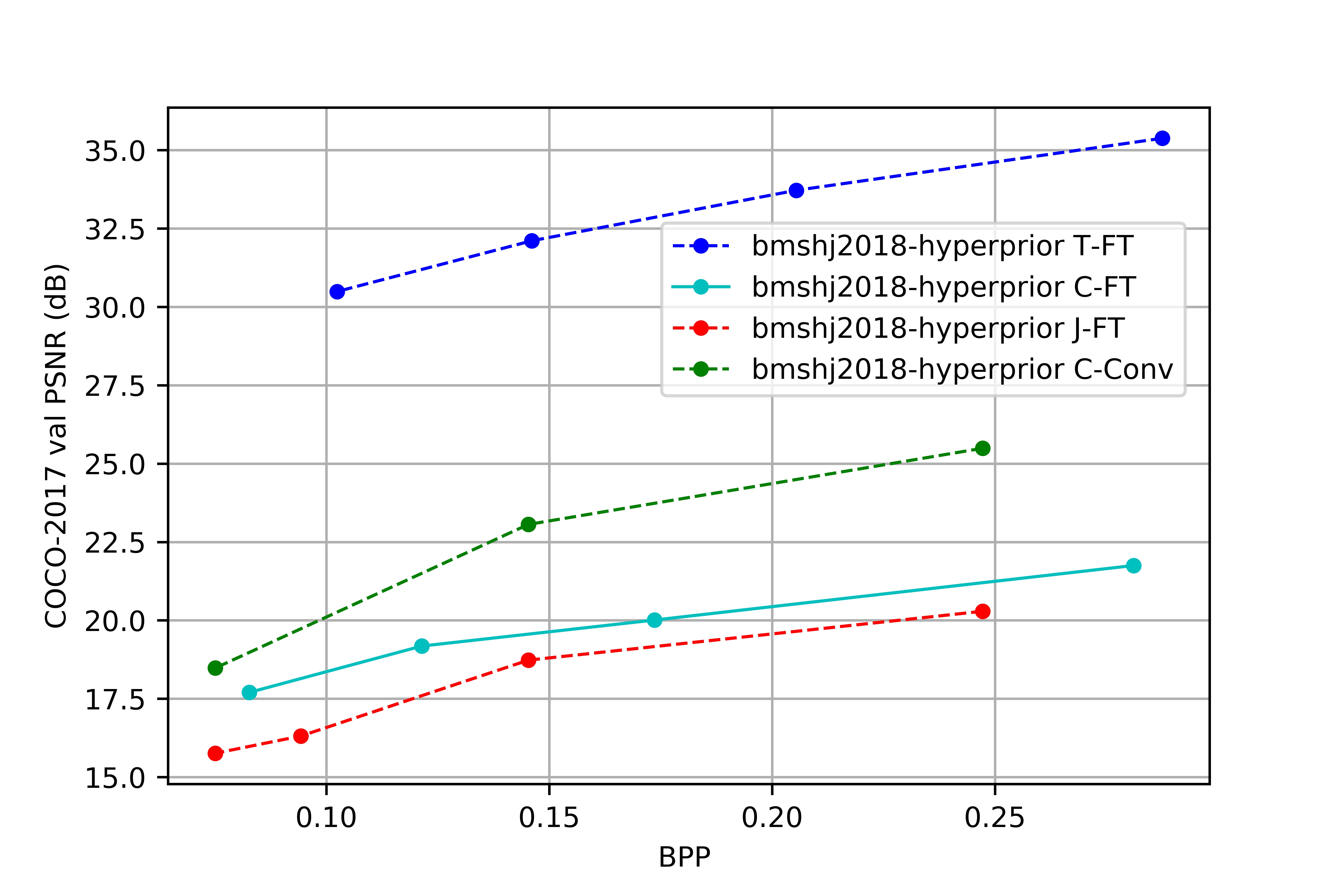}} \subfloat[MS-SSIM]{\includegraphics[width=0.5\linewidth]{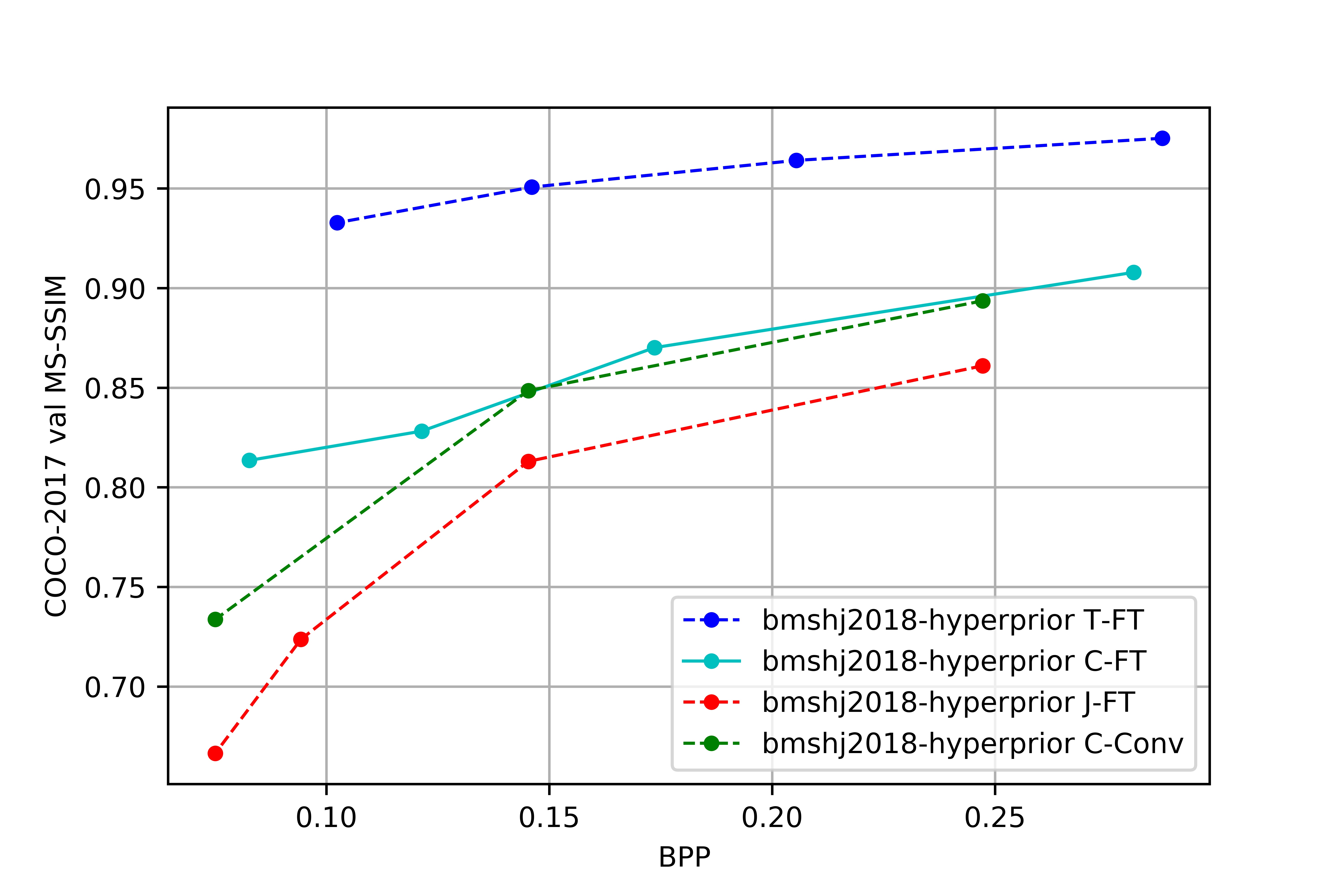}}
	\caption{Rate-distortion characteristics comparison of the proposed methods.}
	\label{fig:supratedistortion}
\end{figure*}

Even though rate-detection-optimized images show better results for secondary tasks, the visual quality of the reconstructed images is significantly lower, compared with rate-distortion or rate-accuracy-distortion\cite{luo2020rate} optimized images, as displayed in Fig.~\ref{fig:jointrecon}. The numerical measures of distortion in PSNR and MS-SSIM can be found in Fig.~\ref{fig:supratedistortion}. Note that the gain of PSNR from the proposed connectors is more significant, compared to MS-SSIM, since the connectors are optimized to minimize the mean square error (MSE).

We selected the object detection task on COCO-2017 as the primary task
considering its high complexity and the variability of the data set.  By
doing so, we expected the end-to-end optimized codec to capture a fair amount of
prior information that is common to vision based learning tasks.  Furthermore
with the intention of preserving the features that is backbone dependent (if
there is any), we used models with ResNet-50 backbone for all 3 tasks in the
experiments.
\section{Comparison of the reconstruction quality of sample images from ImageNet-1K and Cityscapes}
In this section we present additional reconstructed images to highlight the effects of the proposed connectors. The reported rate values for each sample correspond to the average BPP value for the validation set containing that sample image.
\begin{figure*}[!htbp]
	\setlength{\tabcolsep}{1pt}
	\caption{Reconstruction quality for ImageNet-1K}
	\label{fig:avgpool1}%
	\centering
	\begin{tabular}{rc|c|c}
		Original & J-FT&J-FT Avg pool& J-FT Conv\\	
		\includegraphics[width=0.21\linewidth]{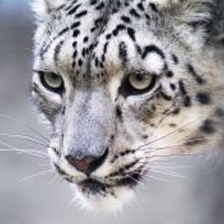}&
		\includegraphics[width=0.21\linewidth]{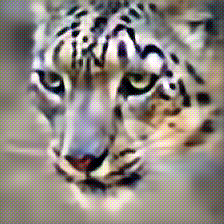}&
		\includegraphics[width=0.21\linewidth]{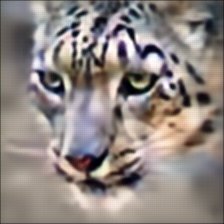}&
		\includegraphics[width=0.21\linewidth]{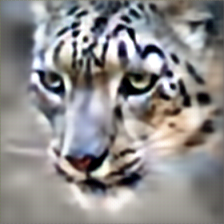}
		\\
		PSNR&15.55&17.36&18.10\\
		BPP&0.1119&0.1119&0.1119\\
		&
		\includegraphics[width=0.21\linewidth]{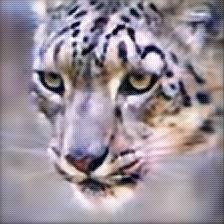}&
		\includegraphics[width=0.21\linewidth]{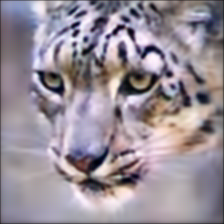}&
		\includegraphics[width=0.21\linewidth]{figures/imagenet/snowLQ4JointConv.png}
		\\
		PSNR&18.19&20.67&21.14\\
		BPP&0.2203&0.2203&0.2203\\
\end{tabular}\end{figure*}

\begin{figure*}[!htbp]
	\setlength{\tabcolsep}{1pt}
	\caption{Reconstruction quality for ImageNet-1K}
	\label{fig:avgpool2}%
	\centering
	\begin{tabular}{rc|c|c}
		Original & J-FT&J-FT Avg pool& J-FT Conv\\	
		\includegraphics[width=0.21\linewidth]{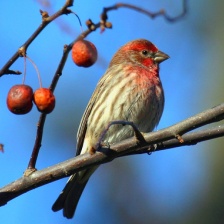}&
		\includegraphics[width=0.21\linewidth]{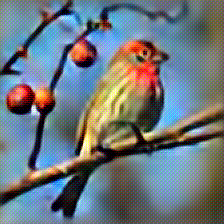}&
		\includegraphics[width=0.21\linewidth]{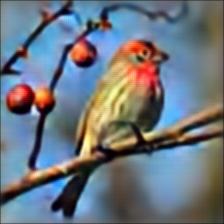}&
		\includegraphics[width=0.21\linewidth]{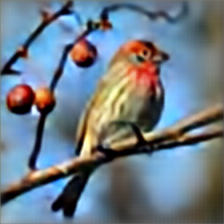}
		\\
		PSNR&15.55&17.36&18.10\\
		BPP&0.1119&0.1119&0.1119\\
		&
		\includegraphics[width=0.21\linewidth]{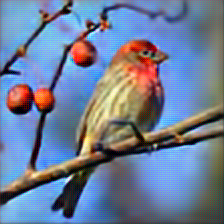}&
		\includegraphics[width=0.21\linewidth]{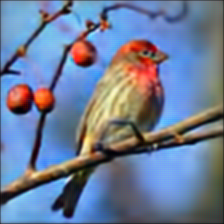}&
		\includegraphics[width=0.21\linewidth]{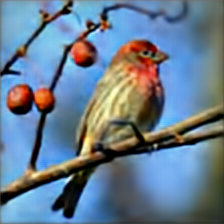}
		\\
		PSNR&18.19&20.67&21.14\\
		BPP&0.2203&0.2203&0.2203\\
\end{tabular}\end{figure*}
\begin{figure*}[!htbp]
	\setlength{\tabcolsep}{1pt}
	\caption{Reconstruction quality for ImageNet-1K}
	\label{fig:avgpool3}%
	\centering
	\begin{tabular}{rc|c|c}
		Original & J-FT&J-FT Avg pool& J-FT Conv\\	
		\includegraphics[width=0.21\linewidth]{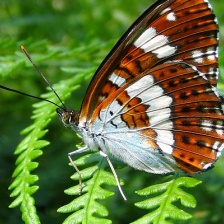}&
		\includegraphics[width=0.21\linewidth]{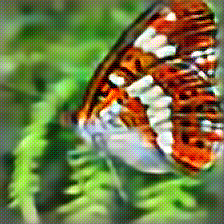}&
		\includegraphics[width=0.21\linewidth]{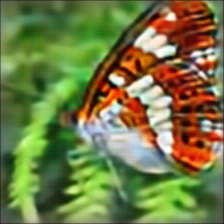}&
		\includegraphics[width=0.21\linewidth]{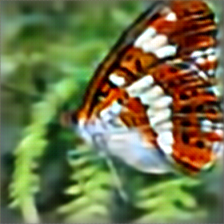}
		\\
		PSNR&15.20&17.11&17.30\\
		BPP&0.1119&0.1119&0.1119\\
		&
		\includegraphics[width=0.21\linewidth]{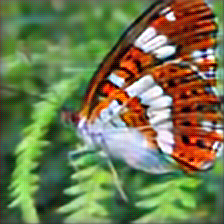}&
		\includegraphics[width=0.21\linewidth]{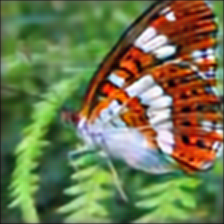}&
		\includegraphics[width=0.21\linewidth]{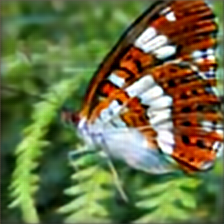}
		\\
		PSNR&16.51&18.02&19.54\\
		BPP&0.2203&0.2203&0.2203\\
	\end{tabular}
\end{figure*}
\begin{figure*}[!htbp]
	\setlength{\tabcolsep}{1pt}
	\caption{Reconstruction quality for Cityscapes}
	\label{fig:avgpool4}%
	\centering
	\begin{tabular}{rc|c|c}
		Original & J-FT&J-FT Avg pool& J-FT Conv\\	
		\includegraphics[width=0.21\linewidth]{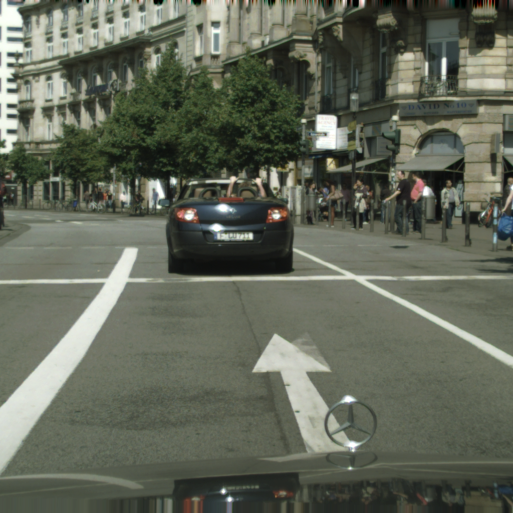}&
		\includegraphics[width=0.21\linewidth]{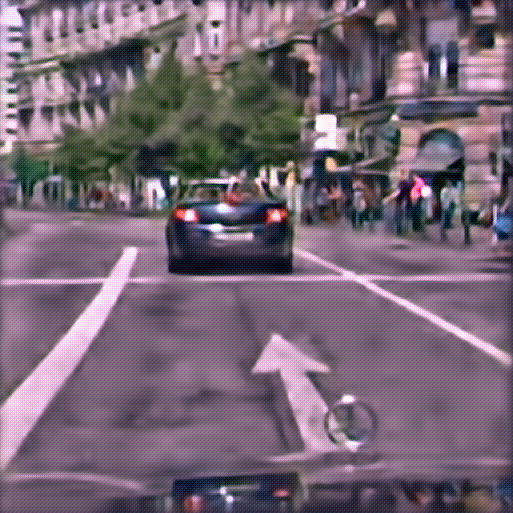}&
		\includegraphics[width=0.21\linewidth]{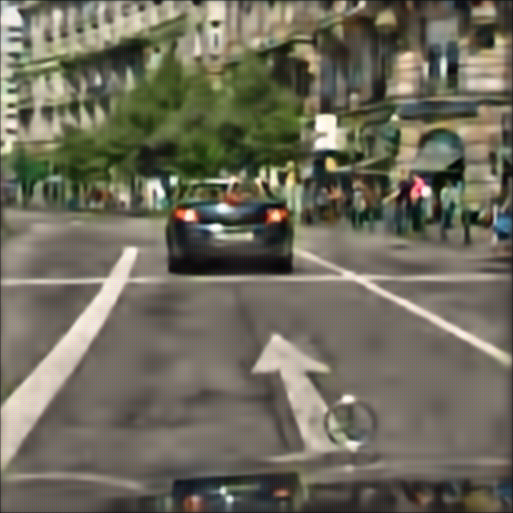}&
		\includegraphics[width=0.21\linewidth]{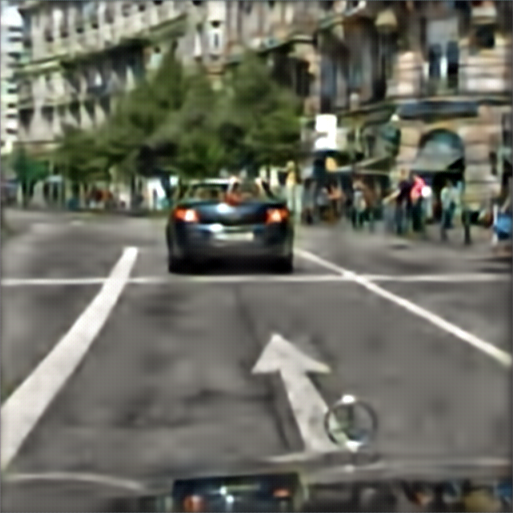}
		\\
		PSNR&17.93&21.34&21.86\\
		BPP&0.0687&0.0687&0.0687\\
		&
		\includegraphics[width=0.21\linewidth]{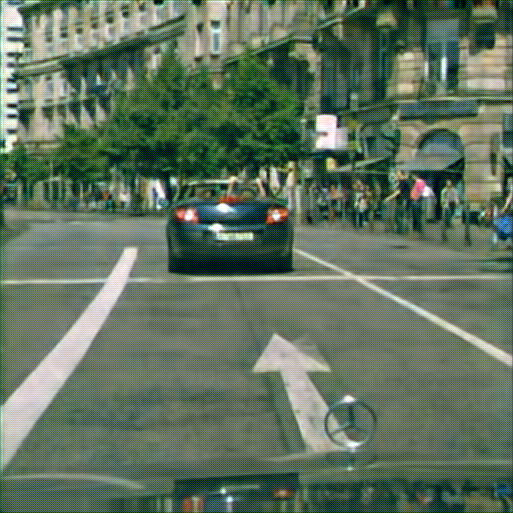}&
		\includegraphics[width=0.21\linewidth]{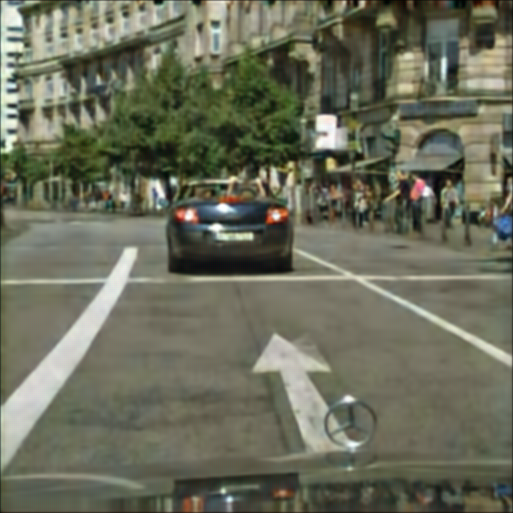}&
		\includegraphics[width=0.21\linewidth]{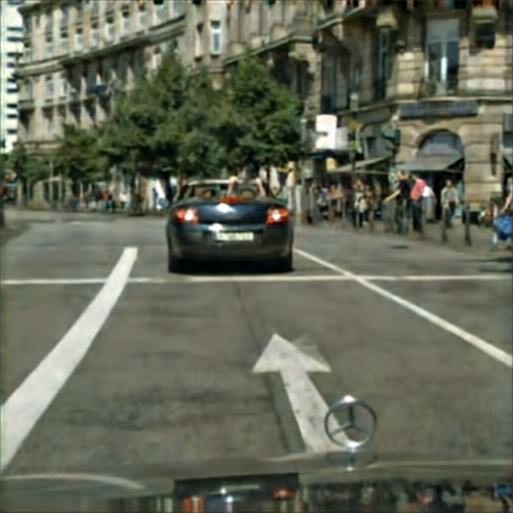}
		\\
		PSNR&20.73&25.03&27.36\\
		BPP&0.2298&0.2298&0.2298\\
	\end{tabular}
\end{figure*}

\bibliographystyle{IEEEtran}
\bibliography{sup_bib}
\end{document}